\pdfoutput=1
\documentclass{llncs}
\usepackage{gensymb}
\usepackage{amssymb}
\usepackage{multirow}
\usepackage{caption}
\usepackage{booktabs}
\usepackage{epsfig}
\usepackage{longtable}
\usepackage{rotating}
\usepackage{amsmath}
\usepackage{array}
\usepackage[bottom]{footmisc}
\usepackage[hyphens]{url} 
\usepackage{color, colortbl}
\usepackage{subfigure}

\usepackage{caption}
\usepackage{epsfig}
\usepackage{longtable}
\usepackage{rotating}
\usepackage{longtable}
\usepackage[bottom]{footmisc}

\begin{document}
\frontmatter          

\title{Automatic Brain Structures Segmentation Using Deep Residual Dilated U-Net}
\author{Hongwei Li, Andrii Zhygallo, and Bjoern Menze\\}
\institute{Technical University of Munich\\
 \email{\{hongwei.li, andrii.zhygallo, bjoern.menze\}@tum.de}\\
}

\maketitle              

\begin{abstract}
Brain image segmentation is used for visualizing and quantifying anatomical structures of the brain. We present an automated approach using 2D deep residual dilated networks which captures rich context information of different tissues for the segmentation of eight brain structures.
The proposed system was evaluated in the MICCAI Brain Segmentation Challenge \footnote{http://mrbrains18.isi.uu.nl/} and ranked 9$^{th}$ out of 22 teams.
We further compared the method with traditional U-Net using leave-one-subject-out cross-validation setting on the public dataset.
Experimental results shows that the proposed method outperforms traditional U-Net (i.e. 80.9\% \emph{vs} 78.3\% in averaged Dice score, 4.35mm \emph{vs} 11.59mm in averaged robust Hausdorff distance) and is computationally efficient.

\end{abstract}
\section{Introduction}
Brain MRI segmentation is an important task in many clinical applications. Various approaches for brain analysis rely on accurate segmentation of anatomical regions. For example, it is commonly used for measuring and visualizing different brain structures, for delineating lesions, for analysing brain development, and for characterization of brain disorders such as Alzheimer’s disease, epilepsy, schizophrenia, multiple sclerosis (MS), cancer, and infectious and degenerative diseases.
Manual segmentation is the gold standard for in-vivo images. However, it requires outlining structures slice-by-slice by neuroradiologist, which is highly time-consuming and prone to rater-bias. Therefore, there is a need for automated segmentation approaches to provide accuracy close to that of expert raters’ with a high reproducibility.

Early works on segmentation of normal brain structures focus on white matter (WM), gray matter (GM), and cerebrospinal fluid (CSF), which is important for studying early brain developments in infants and quantitative assessment of the brain tissue and intracranial volume in large scale studies. Atlas-based approaches \cite{vrooman2007multi,makropoulos2014automatic}, which match intensity information between an atlas and target images and pattern recognition approaches \cite{moeskops2015automatic}, which classify tissues based on a set of local intensity features, are the classical approaches that have been used for brain tissue segmentation.
The MRBrainS Challenge 2013 \cite{mendrik2015mrbrains} was held to compare state-of-the-art segmentation algorithms on three brain structures in conjunction with the 16$^{th}$ International Conference on Medical Image Computing and Computer Assisted Intervention.
Deep-learning based approaches have shown superior performances to the traditional state-of-art methods on the segmentation of brain stroke lesions, brain white matter lesions and brain tumors \cite{maier2017isles,li2018multi,menze2015multimodal}.

In this paper, we presented a deep-learning based method for segmenting eight brain tissues including cortical gray matter (GM), basal ganglia, WM, white matter lesions/hyperintensities (WMH), CSF, ventricles, cerebellum and brain stem.
Deep dilated residual U-Net was adopted to learn context and texture information of different brain tissues.
Multi-sequence data including T1, T1-IR and FLAIR which captures complementary information of different brain structures.
The proposed 2-D network was more computationally efficient than 3D network and traditional U-Net.
Experimental results showed that the proposed method outperforms traditional U-Net.

\section{Materials}

\subsection{Dataset and Protocols}
\subsubsection{Dataset}
Thirty MRI scans were acquired on a 3.0 T Philips Achieva MR scanner at the University Medical Center Utrecht (Netherlands). The following sequences were acquired and used for the evaluation framework: 3D T1 (TR: 7.9 ms, TE: 4.5 ms), T1-IR (TR: 4416 ms, TE: 15 ms, and TI: 400 ms), and T2- FLAIR (TR: 11000 ms, TE: 125 ms, and TI: 2800 ms). The sequences were aligned by rigid registration using Elastix \cite{klein2010elastix} and bias correction was performed using SPM8.
After registration, the voxel size within all provided sequences (T1, T1-IR, and T2-FLAIR) was 0.96 $\times$ 0.96 $\times$3.00 $mm^3$.
Seven scans with annotations were released as a public training set, and the remaining twenty-three scans were used as hidden testing set. For more details on the method of ranking performance, please find the relevant information on the challenge website.
\subsubsection{Evaluation Metric}
Three types of measures were employed to evaluate the segmentation results.
The Dice coefficient is used to determine the spatial overlap and is defined as:
\begin{equation}
	Dice = \frac{2|G \cap P|}{|G|+ |P|}
	\end{equation}
where G is the reference standard, P is the segmentation result.

The 95th-percentile of the Hausdorff distance is used to determine the distance between the segmentation boundaries.
Hausdorff distance is defined as:
	\begin{equation}
	\emph{$H(G,P)$} = max\{\sup\limits_{x\in G} \inf\limits_{y\in P} d(x,y), \sup\limits_{y\in P} \inf\limits_{x\in G} d(x,y)\}
	\end{equation}
	where \emph{d(x, y)} denotes the distance of \emph{x} and \emph{y}, \emph{sup} denotes the supremum and \emph{inf} for the infimum.

The third measure is the volumetric similarity.
Let $V_{G}$ and $V_{P}$ be the volume of lesion regions in $G$ and $P$ respectively.
	Then the volumetric similarity (VS) in percentage is defined as:
	\begin{equation}
	\emph{VS} = \frac{|V_{G}-V_{P}|}{V_{G}}
	\end{equation}

\section{Methodology}
\subsection{Image Preprocessing}
A patient-wise normalization of the image intensities was performed both during training and testing.
For the scan of each patient, the mean value and standard deviation were calculated based on intensities of all voxels.
Then each image volume was normalized to zero mean and unit standard deviation.
Rotation, shearing, scaling along horizontal direction (x-scaling), and scaling along vertical direction (y-scaling) were employed for data augmentation.
After data augmentation, a four times larger training dataset was obtained.

\subsection{2D Dilated Residual U-Net}
\begin{figure*}[t]
	\begin{center}
		\includegraphics[width=1\linewidth,height=0.5\linewidth]{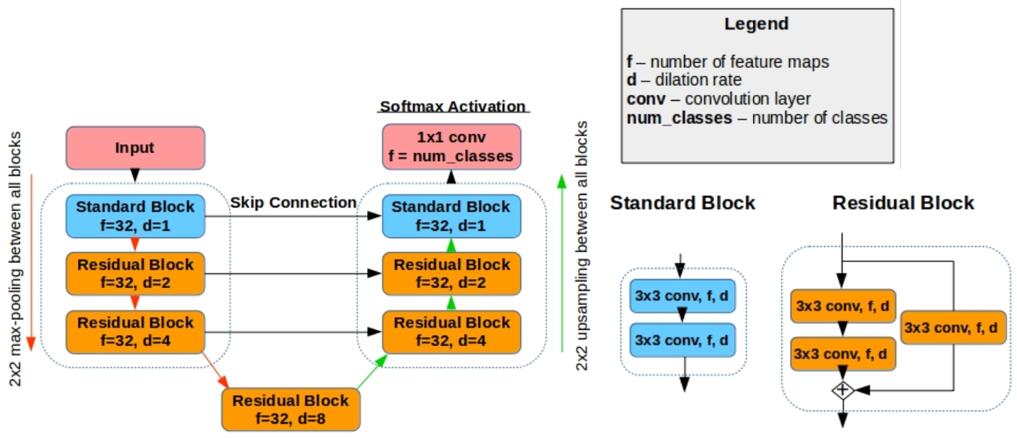}
	\end{center}
	\caption{Details of DRUnet architecture which contains residual blocks with dilated convolutions.}
	\label{fig:drunet} 
\end{figure*}

We used Dilated Residual U-Net (DRUNet), which was originally proposed in \cite{devalla2018drunet} for nerve head tissues segmentation in optical coherence tomography images. DRUNet exploits the inherent advantages of the U-Net skip connections \cite{RFB15a}, residual learning \cite{he2016deep} and dilated convolutions \cite{yu2015multi} to capture rich context information and offer a robust brain structure segmentation with a minimal number of trainable parameters. \newline
DRUNet architecture is presented in Fig.~\ref{fig:drunet}. The model consists of downsampling and upsampling parts. In turn, each part includes one standard block and two residual blocks. Corresponding blocks in downsampling and upsampling parts are connected through skip connections. Convolution layers in both block types have 32 filters of size 3x3. In total the entire network consists of 156$,$105 trainable parameters.
\subsection{Combination of Modalities}
Multi-sequence data including T1-weighted (T1), T1-weighted inversion recovery (T1-IR) and FLAIR which captures complementary information of different brain structures were used for training the network.
In clinical practice, the combination of FLAIR and T1 is beneficial for segmenting white matter lesions while the combination of T1 and T1-IR is helpful for annotating cerebrospinal fluid.
We feed different combinations of modalities for multiple networks.

\subsection{Ensemble Model}
To improve the robustness of our model, an ensemble method was used in the testing stage. Then when given a new testing subject, each subject will be segmented based on the averaged probability maps by the ensemble model.

\begin{figure*}[t]
	\begin{center}
		\includegraphics[width=0.95\linewidth,height=1.1\linewidth]{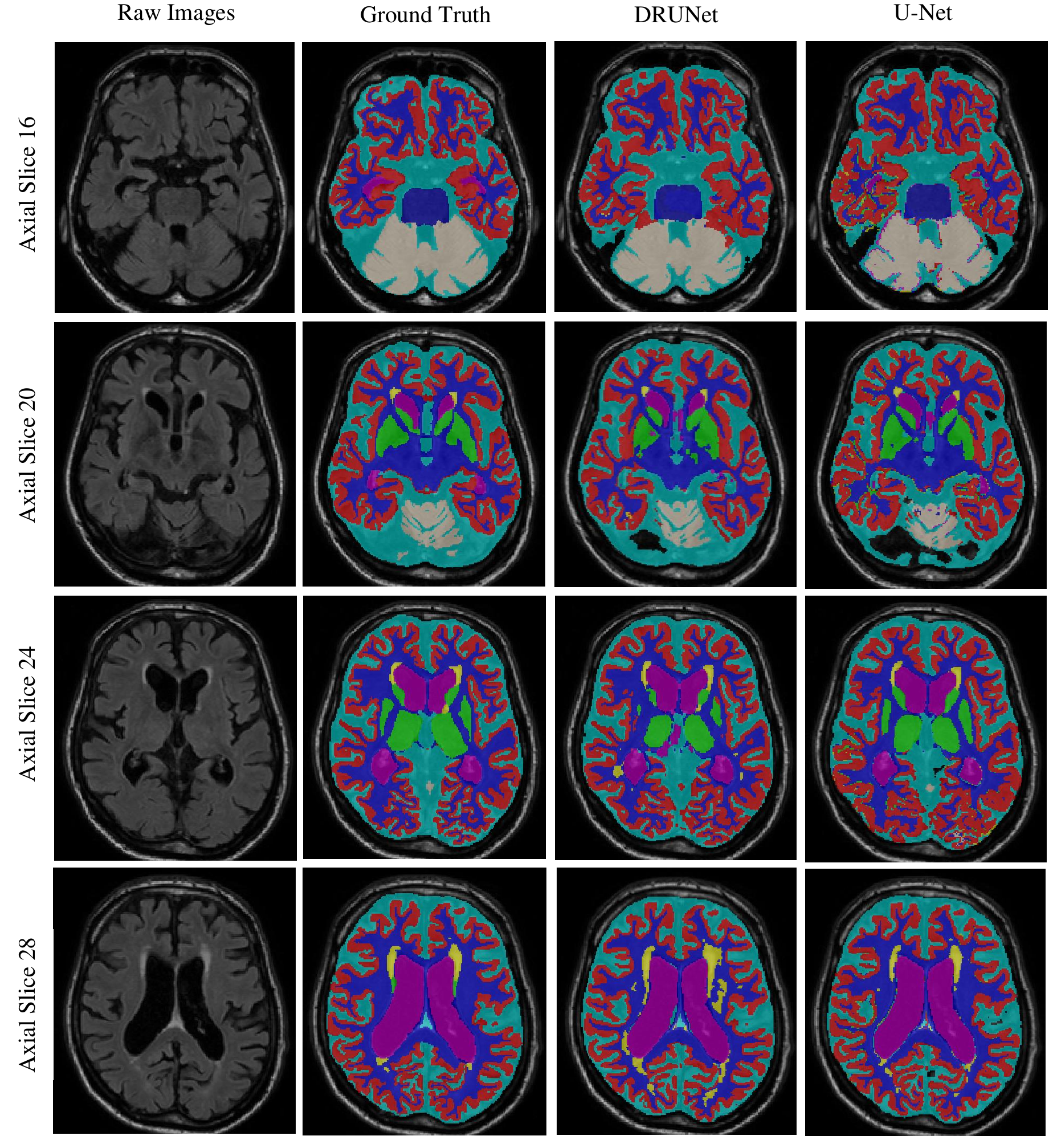}
	\end{center}
	\vspace{0.5cm}
	\caption{Sample segmentation result on \emph{Case 70}. From top to bottom: four axial slices of the same scan. From left to right: FLAIR MR images, the associated ground truth, segmentation result using DRUNet and segmentation result using U-Net. (Best viewed in colour). We can observed from the segmentation result of axial slice 16 that DRUNet achieved better performance on large continuous regions while U-Net generated some isolated false positives. It indicates that the dilated convolution in DRUNet helps to capture context information. On the other hand, for the segmentation of small tissues such as WMHs, DRUNet seems to generate more false positives than U-Net as observed from axial slice 28.}
	\label{fig:samples} \vspace{0.5cm}
\end{figure*}

\subsection{Our Submissions}
\subsubsection{Submission 1}
We used only DRUNet for simultaneously segmenting the ten labels including infarction and pathologies were set to background label during the training of the network.
We generated five DRUNet models with the same architecture but trained with shuffled batches.
Then in testing stage, each subject was segmented based on the averaged probability maps by the ensemble models.
\subsubsection{Submission 2}
We used two Dilated Residual U-Nets (DRUNet) and one traditional U-Net for segmenting different labels.
Since not all the labels were annotated in the same modalities, i.e., white matter lesions were annotated on the FLAIR scan and the outer border of the CSF was segmented using both the T1-weighted scan and the T1-weighted inversion recovery scan, we employed a multi-stage approach to segment different tissues from coarse to fine using different combinations of input modalities.
Firstly, coarse segmentation including eight brain tissues (other labels including infarction and pathologies were set to background label) was performed using FLAIR and T1-weighted modalities by DRUNet (model~1).
Secondly, CSF was independently segmented using T1 and T1-IR modalities by DRUNet (model~2).
Thirdly, since segmentation of white matter lesions is a very challenging task, we used the pre-trained model of the winning method in MICCAI WMH challenge \cite{li2018fully} (model~3) to perform segmentation independently. Finally we fused the multi-stage segmentation results.
Five DRUNet models for model~1 and model~2, respectively, with the same architecture were trained with shuffled batches.

\section{Results}

\subsection{Leave-one-subject-out Evaluation}
To test the generalization performance of our systems across different subjects, we conducted an experiment on the public training datasets (seven subjects) in a leave-one-subject-out setting. Specifically, we used the subject IDs to split the public training dataset into training and validation sets. In each split, we used slices from six subjects for training, and the slices from the remaining subject for testing. This procedure was repeated until all of the subjects are used as testing.
The results were shown in Table. \ref{tab:LOSO}.
There exists significant segmentation difference on subject 4.
We further observed the brain structures of subject 4 and found it was a heathy brain scan without WMHs, infarctions and other lesions.
The reason for the performance difference could be that the models in first submission were trained on 10 labels including infarctions and other lesions while the models in the second submission were trained on 8 main structures excluding two other labels.
When testing on healthy scans, the models trained with 8 main healthy tissues could be more effective since the data distributions among training and testing were similar.


\begin{table*}[t]

	\scriptsize
	\centering
	\caption{Leave-one-subject-out evaluation of our submissions on the public training set containing seven subjects. The averaged Dice score, averaged H95,averaged volume similarity of eight tissues for each subject were shown in the table. The left and right values in each cell were the results of submission~1 and submission~2 respectively. The values in bold indicates the subject on which the two submissions has significant segmentation difference.}
	\begin{tabular}{cccccccc}
		\toprule
		\textbf{Metrics}&Subject~1&~Subject~2&~Subject~3&~Subject~4&~Subject~5&~Subject~6&~Subject~7\\
    	\midrule
		\emph{Dice}&~0.86/0.85&~0.82/0.82&~0.77/0.77&~\textbf{0.73/0.77}&~0.85/0.84&~0.80/0.80&~0.83/0.81 \\
		\emph{H95}&~2.98/2.43&~3.15/2.33&~6.07/6.56&~\textbf{8.25/5.87}&~3.42/2.39&~4.17/6.58&~2.49/8.07 \\
		\emph{VS}&~0.97/0.98&~0.92/0.91&~0.86/0.87&~\textbf{0.82/0.88}&~0.94/0.92&~0.88/0.89&~0.92/0.90 \\

		\bottomrule
	\end{tabular}
\label{tab:LOSO}
\end{table*}

\subsection{Comparison with U-Net}
We further compared the performance of the proposed method (submission~1) with traditional U-Net using the state-of-the-art architecture proposed in \cite{li2018fully}. As shown in Table. \ref{tab:comparison}, generally our approach outperformed traditional U-Net, especially in segmentation of WM and CSF, with an improvement of 8\% and 11\% in Dice score.
WM and CSF are both large structures in brains. We concluded that the use of dilated convolutions is beneficial for capturing the context information of large target.
Furthermore, our model is with much fewer trainable parameters (156$,$105 \emph{vs} 8$,$748$,$609). Thus the training of the network is computationally efficient. The segmentation results of both DRUNet and U-Net on test \emph{case 70} was shown in \ref{fig:samples}.

\begin{table*}[t]

	\scriptsize
	\centering
	\caption{Comparison on each class with traditional U-Net under leave-one-subject-out setting.
The performance on each class was averaged over seven subjects. The values in bold indicated significant improvement over traditional U-Net.}\label{table:Table2}.
	\begin{tabular}{cccccccccc}
		\toprule
		\textbf{Metrics}&~~GM~~&~BG~~&~WM~&~~WMH~&~~CSF~~&~Ventricles&~Cerebellum&~Brain~Stem &\emph{Averaged}\\
    	\midrule
		Dice$_{U-Net}$&~0.83&~0.84&~0.70&~0.79&~0.43&~0.89&~0.9&~0.88&~\emph{0.783} \\
		Dice$_{DRUNet}$&~0.84&~0.85&~\textbf{0.78}&~\textbf{0.81}&~\textbf{0.54}&~0.88&~\textbf{0.92}&~0.85&~\textbf{\emph{0.809}} \\
	    \midrule
		H95$_{U-Net}$&~1.26&~1.8&~43.5&~1.78&~23.09&~3&~15.63&~2.67 &~\emph{11.59} \\
		H95$_{DRUNet}$&~1.29&~1.67&~\textbf{5.82}&~1.61&~\textbf{14.8}&~3.15&~\textbf{2.97}&~3.45&~\textbf{\emph{4.35}} \\
	    \midrule
		VS$_{U-Net}$&~0.95&~0.95&~0.84&~0.93&~\textbf{0.71}&~0.94&~0.96&~0.94&~\emph{90.25}\\
		VS$_{DRUNet}$&0.96&~0.94&~\textbf{0.89}&~0.94&~0.66&~0.93&~0.97&~0.92&~\emph{90.13}\\

		\bottomrule
	\end{tabular}
\label{tab:comparison}
\end{table*}

\subsection{Results on Hidden Testing Cases}
\begin{figure*}[t]
	\begin{center}
		\includegraphics[width=1\linewidth,height=0.7\linewidth]{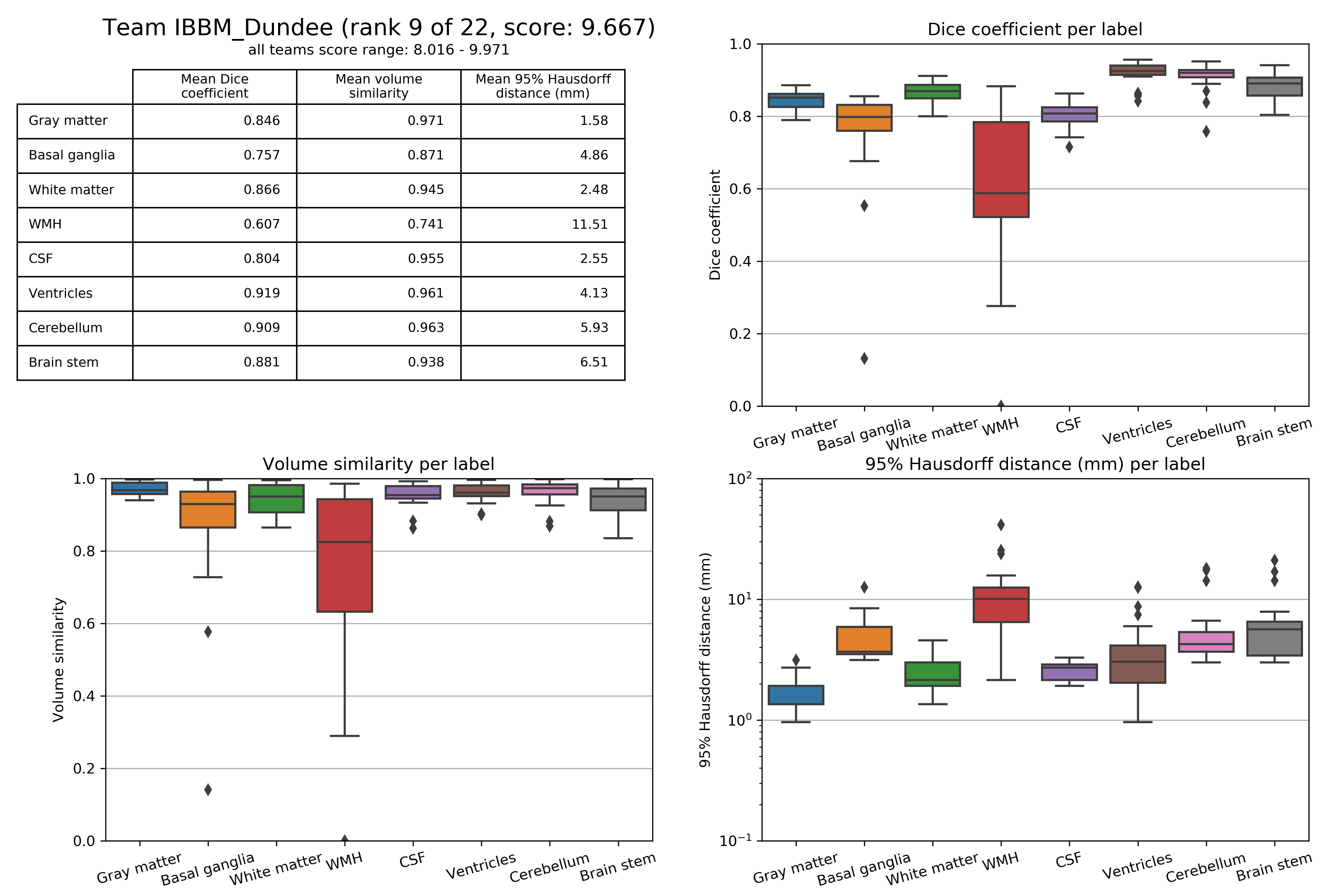}
	\end{center}
	\caption{Result of our first submission on the 23 hidden testing set evaluated by the challenge organizers.
Our method achieved Dice scores of more than 80\% and volume similarity of more 90\% on the major classes while the segmentation performance on WMHs is relatively poor. This is because the WMHs are in small volumes and thus the most difficult structure to be segmented.}
	\label{fig:result_1} 
\end{figure*}

\begin{figure*}[t]
	\begin{center}
		\includegraphics[width=1\linewidth,height=0.7\linewidth]{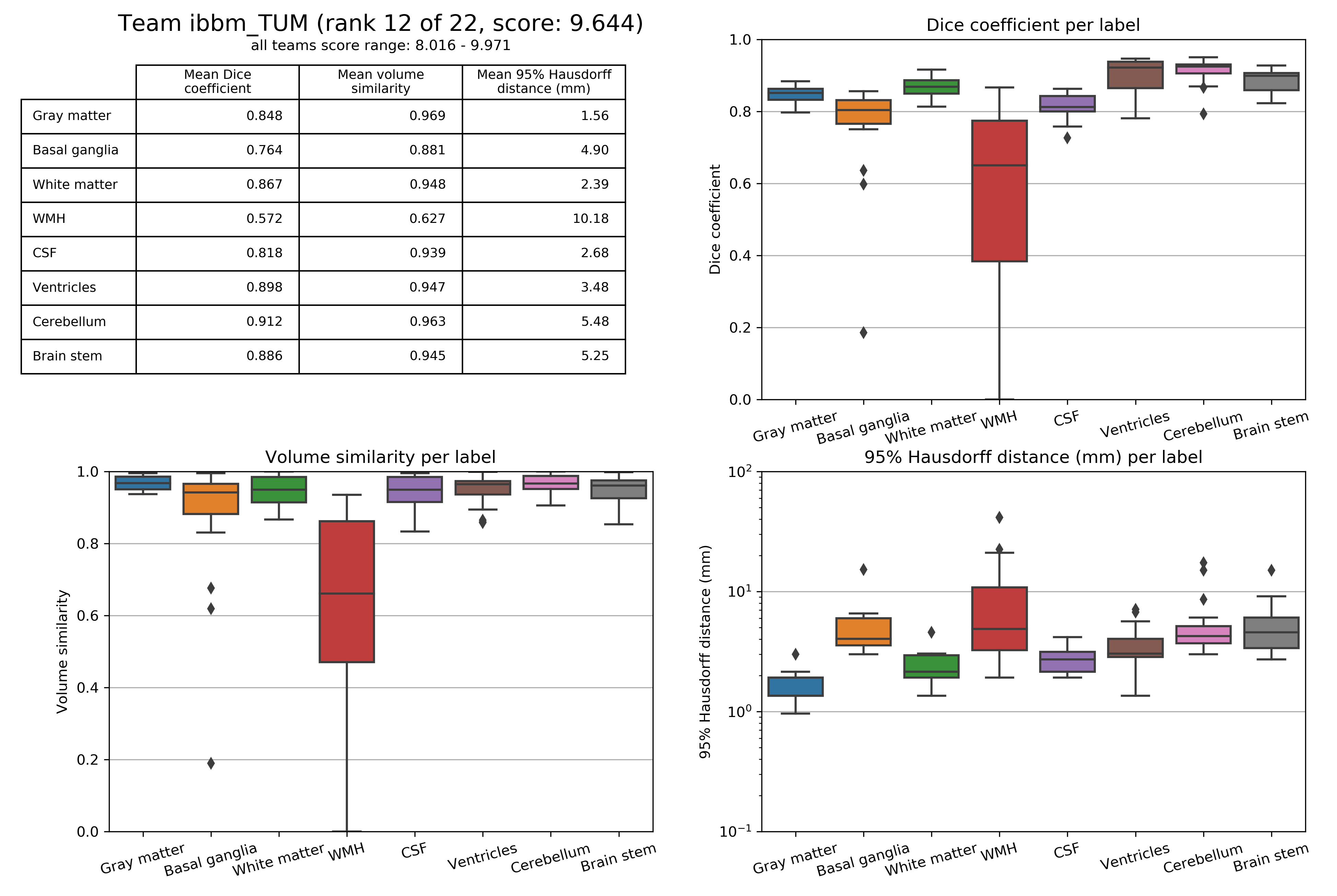}
	\end{center}
	\caption{Result of our second submission on the 23 hidden testing set evaluated by the challenge organizers.
The two submissions achieved comparable performance on major classes except the WMHs. Actually the second submission was designed to improve the segmentation performance of WMHs and integrated the state-of-the-art models from \cite{li2018fully}. There may exist some implementation mistakes in the label fusion stage.}
	\label{fig:result_2} 
\end{figure*}

Our submissions were independently evaluated by the challenge organizer.
Fig. \ref{fig:result_1} and Fig. \ref{fig:result_2} show the box plots of performance on eight labels on 23 testing scans.
Submission~1 and submission 2 ranked $9^{th}$ and $12^{th}$ respectively out of 22 teams.

\bibliographystyle{splncs03}
\bibliography{egbib}

\end{document}